%% file: paper.tex
\documentclass[runningheads]{llncs}
\pdfoutput=1 
\usepackage{graphicx}
\usepackage{amsmath,amssymb,bbm} %
\usepackage{color}
\usepackage{xspace}
\usepackage{lettrine}
\usepackage{array}
\usepackage[width=122mm,left=12mm,paperwidth=146mm,height=193mm,top=12mm,paperheight=217mm]{geometry}

\def\etal{\emph{et al.}\xspace}
\def\ie{\emph{i.e.}\xspace}
\def\eg{\emph{e.g.}\xspace}
\def\vs{\emph{vs}\xspace}
\def\Re{\mathbb{R}}
\def\I{{\cal I}}
\def\one{\mathbbm{1}}

\renewcommand\paragraph[1]{\medskip \noindent \textbf{#1.}}

\newcommand\tobeadded[1]{}

\begin{document}
\pagestyle{headings}
\mainmatter
\def\ECCV16SubNumber{570}  %

\title{Polysemous codes} %

\titlerunning{Polysemous codes}
\authorrunning{Douze, J\'egou \& Perronnin}
\author{Matthijs Douze, Herv\'e J\'egou and Florent Perronnin}
\institute{Facebook AI Research}

\maketitle

\input{abstract.tex}

\input{introduction.tex}

\input{related.tex}

\input{polysemous.tex}

\input{experiments.tex}

\input{graph.tex}

\input{conclusion.tex}

{
\bibliographystyle{splncs03}
\bibliography{egbib,egbibJ}
}
\end{document}

%% file: abstract.tex
\begin{abstract}
This paper considers the problem of approximate nearest neighbor search in the compressed domain. We introduce polysemous codes, which offer both the distance estimation quality of product quantization and the efficient comparison of binary codes with Hamming distance.  Their design is inspired by algorithms introduced in the 90's to construct channel-optimized vector quantizers. At search time, this dual interpretation accelerates the search. Most of the indexed vectors are filtered out with Hamming distance, letting only a fraction of the vectors to be ranked with an asymmetric distance estimator.

The method is complementary with a coarse partitioning of the feature space such as the inverted multi-index. 
This is shown by our experiments performed on several public benchmarks such as the BIGANN dataset comprising one billion vectors, for which we report state-of-the-art results for query times below 0.3\,millisecond per core. Last but not least, our approach allows the approximate computation of the k-NN graph associated with the Yahoo Flickr Creative Commons 100M, described by CNN image descriptors, in less than 8 hours on a single machine.
\end{abstract}

%% file: introduction.tex
\section{Introduction}
\label{sec:introduction} 

\lettrine{N}{earest} neighbor search, or more generally similarity search, has received a sustained attention from different research communities in the last decades. The computer vision community has been especially active on this subject, which is of utmost importance when dealing with very large visual collections. 

While early approximate nearest neighbor (ANN) methods were mainly optimising the trade-off between speed and accuracy,
many recent works~\cite{LCL04,TFW08,JDS11,BL12} put memory requirements as a central criterion for several reasons. 
For instance, due to the memory hierarchy, using less memory means using faster memory: 
disks are slower than the main memory, the main memory is slower than CPU caches, etc. 
Accessing the memory may be the bottleneck of the search. 
Therefore algorithms using compact codes are likely to offer a better efficiency than those relying on full vectors. For these reasons, we focus on ANN search with compact codes, which are able to make search in vector sets comprising as much as one billion vectors on a single machine. 

We distinguish two separate lines of research in ANN with compact codes. 
The first class of methods proposes to map the original vectors to the Hamming hypercube~\cite{LCL04,WTF09,KG09,WKC12}. The resulting bit-vectors are efficiently compared with the Hamming distance thanks to optimized low-level processor instructions such as \texttt{xor} and \texttt{popcnt}, available both on CPUs and GPUs. 
Another increasingly popular approach~\cite{JDS11,NF13,GHKS13,BL14,ZDW14,ZQTW15} is to adopt a quantization point of view to achieve a better distance estimation for a given code size. 
While these two classes of approaches are often seen as contenders, they both have their advantages and drawbacks. Binary codes offer a faster elementary distance computation and do not need external meta-data once the codes are produced. In contrast, quantization-based approaches achieve better memory/accuracy operating points. 

\begin{figure}[t]
~ \hfill
Regular PQ codes
\hfill
~
\hfill
Polysemous PQ codes
\hfill 
~

~
\hfill
\includegraphics[height=0.32\linewidth,trim = 5cm 3cm 5cm 3.5cm,clip]{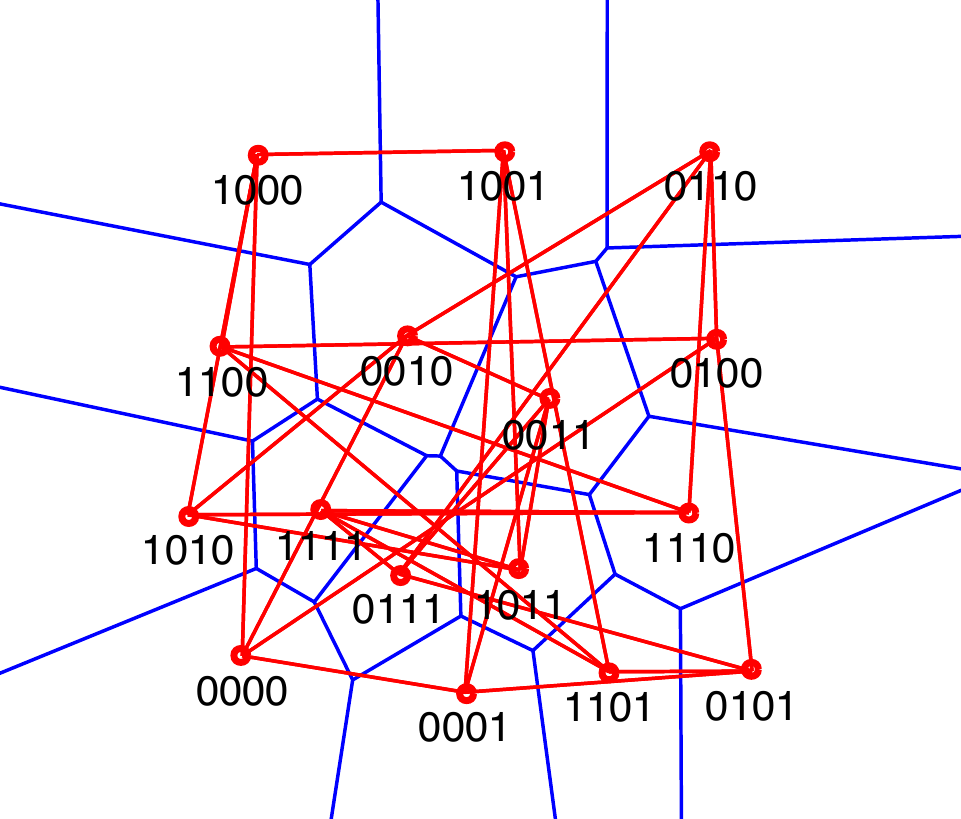} 
\hfill
\includegraphics[height=0.32\linewidth,trim = 5cm 3cm 5cm 3.5cm,clip]{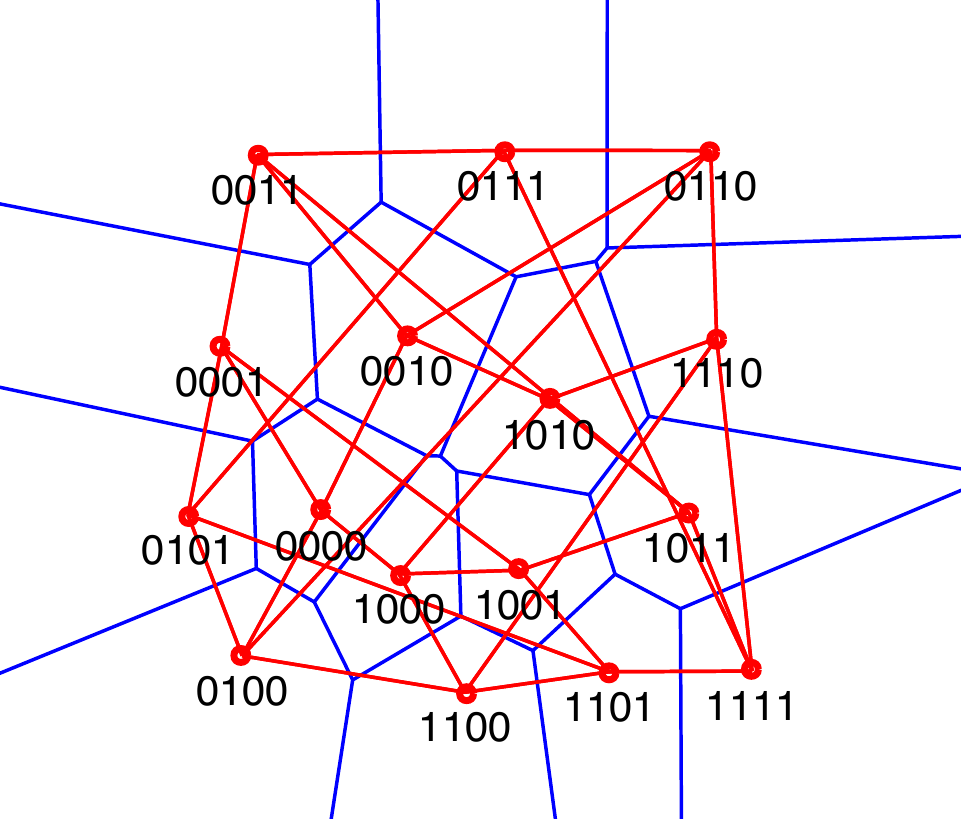} 
\hfill
~
\caption{
Polysemous codes are compact representations of vectors that can be compared either with product quantization (222M distance evaluations per second per core for 8-byte codes) or as binary codes (1.19G distances per second). 
To obtain this property, we optimize the assignment of quantization indexes to bits such that closest centroids have a small Hamming distance. The figure shows k-means centroids (learned on points uniformly drawn in $[0,1]\times[0,1]$) and their corresponding binary representations. %
Observe how the codes differing by one bit (connected by red segments in the figure) generally correspond to close centroids after our optimization (\emph{right}), which is not the case for standard PQ codes (\emph{left}). %
\label{fig:trailer}}
\end{figure}

The polysemous codes introduced in this paper offer the best of both worlds.
They can be compared either with binary codes, which is especially useful in a filtering step, or with the asymmetric distance estimator of product quantization  approaches. The key aspect to attain this dual interpretation is the learning procedure. Our approach is inspired by  works on channel-optimized vector quantization~\cite{F90}. We start by training a product quantizer~\cite{JDS11}. 
We then optimize the so-called \emph{index assignment} of the centroids to binary codes. 
In other terms, we re-order the numeration of the centroids such that distances between similar centroids are small in the Hamming space, 
as illustrated in Figure~\ref{fig:trailer}. 

As a result, our method is almost on par both with quantization-based methods in terms of accuracy and binary methods with respect to search efficiency. 
When combining this approach with a complementary approach such as the inverted multi-index~\cite{BL12}, we outperform the state of the art by a large margin, 
as shown by our experiments carried out on several large public benchmarks. Interestingly, the high efficiency of our approach offers a scalable solution to the all-neighbor problem, \ie, to compute the k-NN graph, for the large image collection Flickr100M described by 4,096 dimensional vectors.

This paper is organized as follows. After briefly reviewing related works on ANN search with compact codes in Section~\ref{sec:related}, Section~\ref{sec:polysemous} describes the design of polysemous codes. The experiments analyzing our approach and comparing to the state-of-the-art are detailed in Section~\ref{sec:experiments}. Finally Section~\ref{sec:graph} illustrates our method on the task of constructing an image graph on a large scale.

%% file: related.tex
\section{Related work: ANN with compact codes}
\label{sec:related}
The literature on efficient search with compact codes is vast and we refer the reader to two recent surveys~\cite{WSSJ14,WLKC15} for extensive references on this subject. 
In this section, we present only a few popular approaches.
\smallskip

\paragraph{Compact binary codes} 
Locality-Sensitive hashing~\cite{IM98,GIM99,C02} is a pioneering binary encoding technique. 
Charikar~\cite{C02} shows under some assumptions that the Hamming distance is statistically related to the cosine similarity 
(equivalently the Euclidean distance for normalized vectors). 
Brute-force comparison of binary hashes has been seen as a viable option for efficient image search with memory constraints~\cite{LCL04}, 
which was popularized by subsequent works evidencing the scalability of this approach to million-sized image collections~\cite{TFF08}. 
Additionally, Norouzi and Fleet have proposed an algorithm to speed-up the search in this Hamming space~\cite{NPF12}. 
Many variants have been subsequently proposed, such as spectral hashing~\cite{WTF09} or ITQ~\cite{GL11} --
see also~\cite{RL10,XWLZLY11,WKC12} for representative works. 
Related to our work, the k-means hashing method~\cite{HeCVPR13} first produces a vector quantizer where the produced codes are compared with the Hamming distance. %

\smallskip

\paragraph{Quantization-based codes} Several works have primarily focused on optimizing the trade-off between memory and distance estimation. 
In particular, it is shown that vector quantizers~\cite{JDS11} satisfying the Lloyd conditions~\cite{GN98} 
offer statistical guarantees on the square Euclidean distance estimator, which is bounded in expectation by the quantizer squared loss. 
These quantization-based methods include product quantization (PQ) \cite{JDS11} and its optimized versions ``optimized product quantization'' \cite{GHKS13} and ``Cartesian $k$-means'' \cite{NF13}. 

These approaches are effective for approximate search within large collections of visual descriptors. Subsequent works~\cite{BL14,ZDW14,ZQTW15} have pushed possible memory/efficiency trade-off by adopting a more general point of view, such as ``Additive quantization'' \cite{BL14}, which provides an excellent approximation and search performance, yet obtained with a much higher computational encoding cost~\cite{VL15}. In between PQ and this general formulation, good trade-offs are achieved by residual quantizers~\cite{BRN96,JH82}, 
which are routinely used in the non-exhaustive PQ variant~\cite{JDS11} to reduce the quantization loss by encoding the residual error vector instead of the original vector, but also as a coding strategy on its own~\cite{CGW10,MHL14,AYWHG15}. 
\smallskip

\paragraph{Non-exhaustive search} 
The aforementioned methods for ANN search limit the memory usage per indexed vector and provide a distance estimator that is faster to compute than the exact distance. However, the search is still exhaustive in the sense that the query is compared to all database elements. For billion-sized collections, reading the codes in memory is a severe constraint leading to search times in the order of a second, typically. 
The limitation imposed by this memory bottleneck has led to two-stage approaches~\cite{JDS08,JDS11,AINR14,KA14}, 
in which the feature space is first partitioned through hashing or clustering.
Practically, an inverted list storing identifiers and corresponding compact codes is stored for each region. 
At query time, the distance is estimated only for the codes associated with a subset of regions~\cite{LJWCL07,JDS11,BL12}. 
It is also possible to use multiple partitions as in early LSH papers, as done in joint inverted indexing~\cite{XHWS13}. 
These solutions however require several indexing structures %
and are therefore not competitive in terms of memory usage.  
Various partitioning methods have been proposed for the coarse level~\cite{PJA10,BL12}. In particular, the inverted multi-index uses product quantization both to define the coarse level and for coding residual vectors. This strategy offers state-of-the-art performance when further combined with a re-ranking strategy based on codes~\cite{JTDA11}.

\paragraph{Motivation: Binary codes versus quantization-based approaches}
The Hamming distance is significantly faster to evaluate than the distance estimator based on table look-ups involved in quantization methods\footnote{A recent method~\cite{AndreVLDB15} reduces the scanning by employing a lower-bounding look-up table stored in SIMD registers, however this method is efficient only on long inverted lists, which makes it sub-optimal in the usual setting.}. 
From our measurements, the acceleration factor is typically
between $4\times$ and $7\times$, depending on the code length. 
However, binary methods suffer limitations imposed by the Hamming space. 
First, the number of possible distances is at most $d$+1, where $d$ is the binary vector length. 
This problem is partially solved by asymmetric variants of LSH~\cite{DCL08,JJG11,GP11}, 
whose estimations use compact codes for database vectors but not on the query side. 
Yet such asymmetric measures require look-ups, like the methods derived from product quantization, and are therefore more expensive to evaluate than the Hamming distance. 
On the other hand, quantization-based methods offer a better memory/accuracy compromise, which is expected since binarization is a particular case of quantization.

Binary and quantization-based codes have their own advantages and drawbacks.
While the literature usually presents binary and quantized-based codes as concurrent methods, the next section introduces a method that benefits from the advantages of both classes of methods.

%% file: polysemous.tex
\def \dist {\mathrm{dist}}
\def \rank {\mathrm{rank}}

\section{Polysemous codes}
\label{sec:polysemous}

We, now, introduce our strategy to take advantage of the fast computation of Hamming distances while offering the estimation accuracy of quantization-based methods. 
The main idea is to learn a regular product quantizer~\cite{JDS11}, 
and then to optimize the assignment of centroid indexes to binary codes such that the Hamming distance approximates the inter-centroid distance. 
In this section, we first describe the objective functions optimized to achieve this property, and then describe the optimization algorithm. 

Note that for a product quantizer, one typically optimizes separately each of the constituent sub-quantizers.
Therefore, in what follows, we have one objective function (and optimization process) per sub-quantizer.

\subsection{Objective functions}
\label{sec:objective}

We consider two objective functions: one that minimizes a loss based on the distance estimator and 
one that minimizes a ranking loss.

\paragraph{Notation} 
A quantizer is usually described by its set of centroids.
Let $\I$ be the set of centroid indexes: $\I = \{0, 1, \ldots, 2^d-1\}$
and $d=8$ if each (sub-)quantizer encodes the original vectors on one byte as is standard practice.
Let $c_i$ be the reproduction value associated with centroid $i$.
Let $d: \Re^D \times \Re^D \rightarrow \Re^+$ be a distance between centroids, for instance the Euclidean distance.
Let $\pi: \I \rightarrow \{0,1\}^d$ denote a bijective function that maps each centroid index to a different vertex of the unit hypercube.
Finally let $h: \{0,1\}^d \times \{0,1\}^d \rightarrow \Re^+$ be the Hamming distance between two $d$-dimensional binary representations. 

\paragraph{Distance estimator loss}
One possible objective if to find the bijective map $\pi$ such that the distance $d(c_i,c_j)$ between two centroids 
is approximated by the Hamming distance $h(\pi(i), \pi(j))$ between the two corresponding binary codes:
\begin{equation}
\label{eqn:del}
\pi^* = \arg \min_\pi \sum_{i\in \I, j \in \I} \left[ h(\pi(i),\pi(j)) - f(d(c_i,c_j)) \right]^2
\end{equation}
where $f:\Re \rightarrow \Re$ is a monotonously increasing function that maps the distance $d(c_i,c_j)$
between codewords into a range comparable to Hamming distances.
In practice, we choose for $f$ a simple linear mapping. 
This choice is motivated by the following observations. 
The Hamming distance between two binary vectors randomly drawn from $\{0,1\}^d$
follows a binomial distribution with mean $d/2$ and variance $d/4$.
Assuming that the distribution of distances $d(c_i,c_j)$ 
can be approximated by a Gaussian distribution --
which is a good approximation of the binomial --
with mean $\mu$ and standard deviation sigma $\sigma$,
we can force the two distributions to have the same mean and variance.
This yields:
\begin{equation}
f(x) = \frac{\sqrt{d}}{2\sigma} (x-\mu) + \frac{d}{2}
\end{equation}
where $\mu$ and $\sigma$ are measured empirically. 

As, in the context of k-NN, it is more important to approximate small distances than large ones, 
we found out in practice that it is beneficial to weight
the distances in the objective function (\ref{eqn:del}).
This leads to a weighted objective:
\begin{equation}
\label{eqn:wdel}
\pi^* = \arg \min_\pi \sum_{i\in \I, j \in \I} w(f(d(c_i,c_j))) \left[ h(\pi(i),\pi(j)) - f(d(c_i,c_j)) \right]^2.
\end{equation}
We choose a function $w: \Re \rightarrow \Re$ of the form $w(u) = \alpha^{u}$ with $\alpha<1$.
In our experiments, we set $\alpha=1/2$ but we found out that values of $\alpha$ in the range $[0.2, 0.6]$
yielded similar results. 

\paragraph{Ranking loss}
In the context of k-NN search, we are interested in finding a bijective map $\pi$ that preserves the ranking of codewords.
For this purpose, we adopt an Information Retrieval perspective.
Let $(i,j)$ be a pair of codewords such that $i$ is assumed to be a ``query'' 
and $j$ is assumed to be ``relevant'' to $i$.
We will later discuss the choice of $(\mbox{query},\mbox{relevant})$ pairs.
We take as negatives for query $i$ the codewords $k$ such that $d(c_i,c_j)<d(c_i,c_k)$.
The loss for pair $(i,j)$ may  be defined as:
\begin{equation}
r_\pi(i,j) = \sum_{k \in \I} \one \left[d(c_i,c_j) < d(c_i,c_k)\right] \one\left[h(\pi(i),\pi(j)) > h(\pi(i),\pi(k))\right]
\end{equation}
where $\one[u] = 1$ if $u$ is true, 0 otherwise.
It measures how many codewords $k$ are closer to $i$ than $j$ according to the Hamming distance
while, $i$ is closer to $j$ than $k$ according to the distance between centroids.
We note that the previous loss measures the number of correctly ranked pairs which is closely related to Kendall's tau coefficient.

An issue with the loss $r_\pi(i,j)$ is that it gives the same weight to the top of the list as to the bottom.
However, in ranking problems it is desirable to give more weight to errors occurring in the top ranks.
Therefore, we do not use directly the loss $r_\pi(i,j)$ for the pair $(i,j)$, but adopt instead a loss that increases sublinearly with $r_\pi(i,j)$.
More specifically, we follow~\cite{UBG09} and introduce a monotonously decreasing sequence $\alpha_i$ as well as the sequence
$\ell_j = \sum_{i=1}^j \alpha_i$, which increases sublinearly with $j$.
We define the weighted loss for pair $(i,j)$ as
$\ell_{r_\pi(i,j)}$.

A subsequent question is how to choose pairs $(i,j)$. %
One possibility would be to choose $j$ among the $k$-NNs of $i$, in which case we would optimize
\begin{equation}
\pi^* = \arg \min_\pi \sum_{i \in \I} \sum_{j \in k-\mbox{NN}(i)} \ell_{r_\pi(i,j)}. 
\end{equation}
An issue with this approach is that it requires choosing an arbitrary length $k$ for the NN list.
An alternative is to consider all $j \neq i$ as being potentially ``relevant'' 
to $i$ but to downweight the contribution of those $j$'s which are further away from $i$.
In such a case, we optimize
\begin{equation}
\label{eqn:wrnkobj}
\pi^*= \arg \min_\pi \sum_{i \in I,j \in \I} \alpha_{r(i,j)} \ell_{r_\pi(i,j)},
\end{equation}
where we recall that $\alpha_i$ is a decreasing sequence and $r(i,j)$ is the rank of $j$ in the ordered list of neighbors to $i$:
\begin{equation}
r(i,j) = \sum_{k \in \I} \one\left[d(c_i,c_j)<d(c_i,c_k) \right]
\end{equation}
In all our ranking experiments, we use Equation \eqref{eqn:wrnkobj} and choose $\alpha_i = 1/i$ (following~\cite{UBG09}).

\subsection{Optimization}
\label{sec:optimization}

The aforementioned objective functions aim at finding a bijective map $\pi$, 
or equivalently another numeration of the set of PQ centroids, that would assign similar binary codes to neighbouring centroids. 

This problem is similar to that of channel optimized vector quantization~\cite{ZG90,F90,FV91}, 
for which researchers have designed quantizers such that the corruption of a bit by the channel impacts the reconstruction as little as possible. 
This is a discrete optimisation problem that can not be relaxed, and for which we can only target a local minimum, as the set of possible bijective maps is huge. 
In the coding literature, such index assignment problems were first optimised in a greedy manner, for instance by using the binary switching algorithm~\cite{ZG90}. 
Starting from an initial index assignment, at each iteration, this algorithm tests all possible bit swaps (\ie, $d$), and keeps the one providing the best update of the objective function. As shown by Farvardin~\cite{F90}, this strategy rapidly gets trapped in a poor local minimum. 
To our knowledge, the best approach to index assignment problems is to employ simulated annealing to carry out the optimization. 
This choice was shown to be significantly better~\cite{F90} than previous greedy approaches. 
The algorithm aims at optimizing a loss $L(\pi)$ that depends on the bijective mapping $\pi$
defined as a table of size $2^d$. It proceeds as follows

\newcommand{\IND}{\hspace*{2em}}

\begin{enumerate}
\item Initialize 
\item \IND current solution $\pi := [0, ...., 2^d-1]$
\item \IND temperature $t := t_0$
\item Iterate $N_\mathrm{iter}$ times:
\item \IND draw $i,j \in \I, i\ne j$ at random
\item \IND $\pi':=\pi$, with entries $i$ and $j$ swapped 
\item \IND compute the cost update $\Delta C := L(\pi') - L(\pi)$
\item \IND if $\Delta C < 0$ or at random with probability $t$: %
\item \IND \IND accept the new solution: $\pi:=\pi'$
\item \IND $t:=t \times t_\mathrm{decay}$
\end{enumerate}

The algorithm depends on the number of iterations $N_\mathrm{iter}=500,000$, the initial ``temperature'' $t_0=0.7$ and $t_\mathrm{decay} = 0.9^{1/500}$, ie. decrease by a factor 0.9 every 500 iterations. Evaluating the distance estimation loss (resp ranking loss) has a complexity in $\mathcal{O}(2^{2d})$ (resp. $\mathcal{O}(2^{3d})$). However, computing the cost update incurred by a swap can be implemented in $\mathcal{O}(2^d)$ (resp. $\mathcal{O}(2^{2d})$).

\begin{figure}[t]
\includegraphics[height=0.33\linewidth]{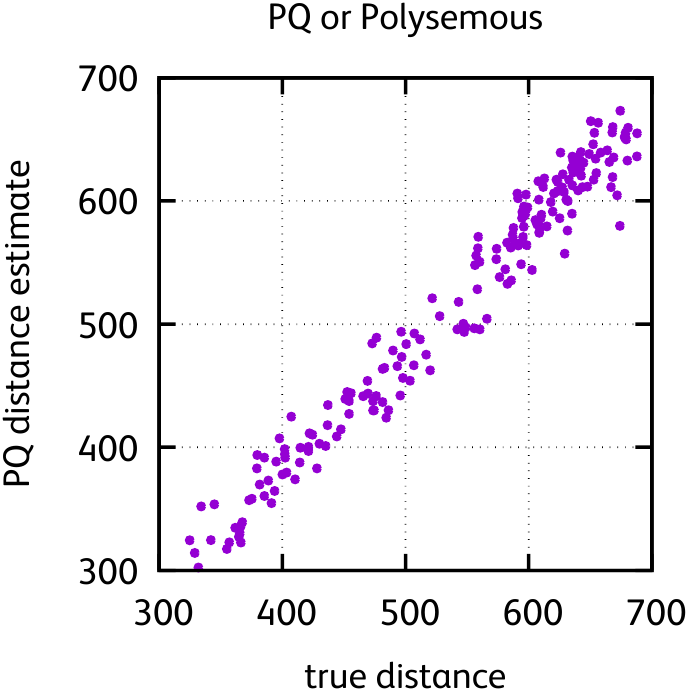}
\includegraphics[height=0.33\linewidth]{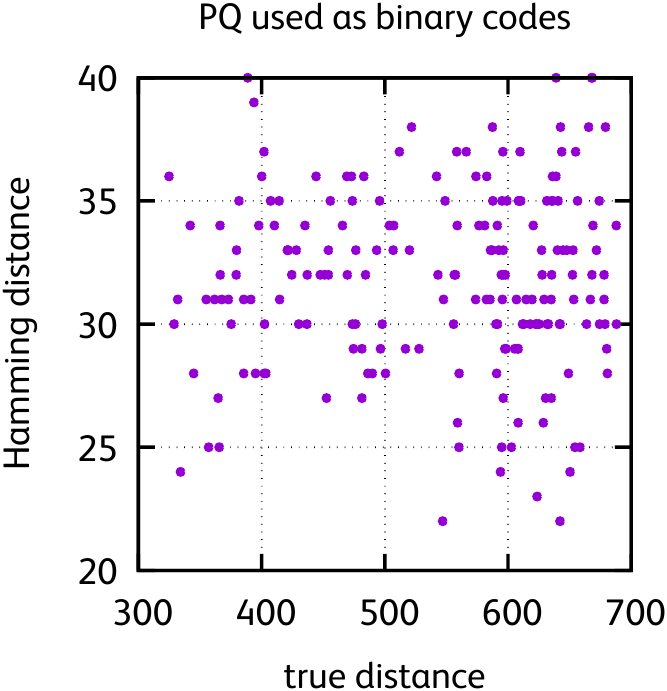}
\includegraphics[height=0.33\linewidth]{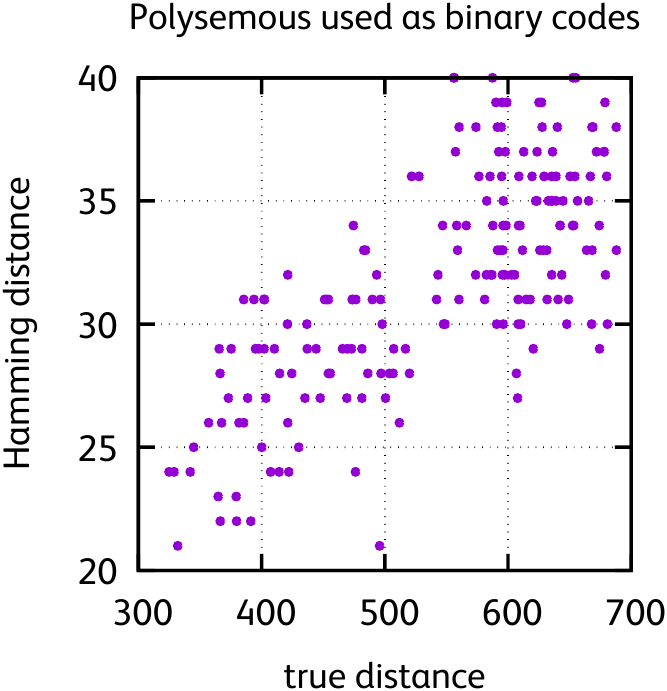}
\caption{\textit{Left:} True distances \vs distance estimates with PQ codes. 
\textit{Middle:} True distances \vs Hamming distances before polysemous optimization. 
\textit{Right:} True distances \vs Hamming distances after polysemous optimization. 
The binary comparison with Polysemous is much more discriminative, while offering the same estimation when being interpreted as PQ codes. 
\label{sec:estimate_vs_real}}
\end{figure}

Figure~\ref{sec:estimate_vs_real} shows on a set of SIFT descriptors that our optimization is effective: 
the comparison of the codes used as binary vectors is much more correlated with the true distance after than before the optimization.

\subsection{Discussion}
\label{sec:discussion}

Although the optimization algorithm is similar to those previously employed in channel-optimized vector quantization, 
our objective functions are significantly different to reflect our application scenario. 
In communications, it is unlikely that many bit errors occur simultaneously, in particular not on a memoryless channel. 
Therefore the objective functions employed in communication focus on small Hamming distances. 
In contrast, for ANN the typical Hamming distances of the neighbors are relatively large. 

We point out that, while the proposed binarized PQ codes offer a competitive performance,
their accuracy is significantly lower than that of PQ.
This suggests a two-step strategy for large-scale search.
Given a query, we first filter out the majority of the database items using the fast Hamming distance on the binarized PQ codes.
We then evaluate the more costly asymmetric distances for the items whose Hamming distance was below a given threshold~$\tau$.

Other strategies could be contemplated for the filtering stage.
One such strategy is to measure how many quantization indexes differ for the product quantizer\footnote{Formally, this quantity is also called Hamming distance, but measured between vector of indexes and not binary vectors.}. 
In other terms, one can filter out vectors if more than a given number of sub-quantizers produce indexes not identical to those of the queries. 
As shown in the experimental section~\ref{sec:experiments}, this method is not as efficient nor precise as the strategy proposed in this section. %

Another such strategy would be to use for the filtering stage a binary encoding technique unrelated to PQ, \eg, ITQ.
The issue is that it would increase the memory requirements of the method as it would involve storing ITQ codes and PQ codes.
In constrat, we only store one polysemous code per database item in the proposed approach -- a must if the emphasis is on storage requirements.

%% file: experiments.tex
\section{Experiments}
\label{sec:experiments}

This section gives an analysis and evaluates our polysemous codes. After introducing the evaluation protocol, we analyze our core approach in different aspects. Then we show that our approach is compatible with the inverted multi-index (IMI) and give a comparison against the state of the art.

\subsection{Evaluation protocol}

We analyze and evaluate our approach with standard benchmarks for ANN as well as a new benchmark that we introduce to evaluate the search quality.

\def\siftM  {S{\small IFT}1M\xspace}
\def\bigann {B{\small IGANN}\xspace}
\def\FYM    {FY{\small CNN}1M\xspace}
\def\FYbig  {FY{\small CNN}90M\xspace}

\begin{description}
\item [\siftM] is a benchmark~\cite{JTDA11} of 128-dimensional SIFT descriptors~\cite{L04}. 
There are one million vectors in the database, plus 100,000 vectors for training and 10,000 query vectors. 
This is a relatively small set that we mainly use for parameter analysis.  
\item [\bigann] is a large-scale benchmark~\cite{JTDA11} widely used for ANN search, also made of SIFT descriptors. It comprises one billion database vectors, 100 million training vectors and 10,000 queries.  
\item [\FYM and \FYbig] are introduced to evaluate the quality of the search with more challenging features.
We leverage the Yahoo Flickr Creative Commons 100M\footnote{Out of which only 95M are available for download today.} image collection~\cite{TSFENPBL15} as follows.
In \FYbig, we split the dataset into three sets: 
90M vectors are to be indexed, 10k vectors serve as queries, 
5M vectors are used for training. 
\FYM uses the same training set and queries, 
but the indexed set is restricted to the first million images for the purpose of analyzing our method. 
We extract convolutional neural networks features~\cite{LBDHHHJ90} following the guidelines of~\cite{BSCL14}:
we compute the activations of the 7\textsuperscript{th} layer of AlexNet~\cite{KSH12}.
This yields 4096-dimensional image descriptors.
Prior to indexing we reduce these descriptors to 256D with PCA and subsequently apply a random rotation~\cite{BSCL14,JDSP10}.
\end{description}

For all datasets, the accuracy is evaluated by recall@$R$. 
This metric measures the fraction of the queries for which the true nearest neighbor is returned within the top $R$ results. 
All reported times are on a single core of a 2.8GHz machine.

\subsection{Analysis of Polysemous codes performance}

\def \pq      {\textsf{PQ}\xspace}
\def \polyd   {\textsf{PolyD}\xspace} 
\def \polyr   {\textsf{PolyR}\xspace} 
\def \adc     {\textsf{ADC}\xspace} 
\def \binary  {\textsf{binary}\xspace} 
\def \disidx  {\textsf{disidx}\xspace} 
\def \hybrid  {\textsf{dual}\xspace} 

We first analyze the performance of polysemous codes. 
Let us introduce notations. We first consider three ways of constructing a product quantizer:
\begin{description}
\item [\pq] is the baseline: we directly use the code produced by the product quantizer, without any optimization of the index assignment;
\item [\polyd] refers to a product quantizer whose index assignment is optimized by minimizing the distance estimator loss introduced in Section~\ref{sec:objective};
\item [\polyr] similarly refers to a PQ optimized with the proposed ranking loss. 
\end{description} 

Once the codebook and index assignment are learned, we consider the following methods to estimate distances based on polysemous codes:
\begin{description}
\item [\adc] is the regular comparison based on an asymmetric distance estimator~\cite{JDS11};
\item [\binary] refers to the bitwise comparison with the Hamming distance when the codes are regarded as bitvectors, like for binary codes (\eg, ITQ);
\item [\disidx]counts how many sub-quantizers give different codes (see Section~\ref{sec:discussion});
\item [\hybrid] refers to the strategy employing both interpretations of polysemous codes: 
the Hamming codes are used to filter-out the database vectors whose distance to the query is above a threshold~$\tau$. 
The indexed vectors satisfying this test are compared with the asymmetric distance estimator.  
\end{description}

\noindent \emph{Note:}
Polysemous codes are primarily PQ codes.
Therefore the performance of polysemous codes and regular PQ is identical when the comparison is independent from the index assignment, which is the case for \adc and \disidx. For instance the combinations \polyd/\adc, \polyr/\adc and \pq/\adc are equivalent both in efficiency and accuracy. 
\smallskip

\newcommand{\std}[1]{}

\def \mysp {\hspace{3pt}}

\begin{table}[t]
{%
~ \hfill \begin{tabular}{|@\mysp l@\mysp|@\mysp c@\mysp c@\mysp|@\mysp c@\mysp c@\mysp | r@\mysp|}
\hline 
 & \multicolumn{2}{c}{\siftM} & \multicolumn{2}{c|}{\FYM } & \\
 & $R$@1 & $R$@100& $R$@1 & $R$@100  & query\,(ms) \\ 
 \hline
\pq/\disidx                      & 0.071 \std{0.0015} & 0.281 \std{0.0043} & 0.031 \std{0.0016} & 0.284 \std{0.0017} &   3.66 \std{0.01} \\

\hline
\pq/\binary                     & 0.036 \std{0.0010} & 0.129 \std{0.0028} & 0.015 \std{0.0004} & 0.124 \std{0.0014} &   1.42 \std{0.01} \\
\polyd/\binary                  & 0.107 \std{0.0019} & 0.503 \std{0.0047} & 0.027 \std{0.0013} & 0.281 \std{0.0017} &   1.45 \std{0.05} \\
\polyr/\binary                  & 0.105 \std{0.0026} & 0.467 \std{0.0019} & 0.022 \std{0.0010} & 0.222 \std{0.0026} &   1.45 \std{0.03} \\
\hline
\pq/\hybrid ($\tau=55$)          &  0.312 \std{0.0033} & 0.507 \std{0.0080} & 0.116 \std{0.0019} & 0.522 \std{0.0047} &   2.59 \std{0.03} \\
\polyd/\hybrid ($\tau=51$)       &  0.441 \std{0.0045} & 0.987 \std{0.0012} & 0.132 \std{0.0014} & 0.804 \std{0.0017} &   2.53 \std{0.05} \\
\polyr/\hybrid ($\tau=53$)       &  0.439 \std{0.0060} & 0.960 \std{0.0020} & 0.130 \std{0.0013} & 0.745 \std{0.0026} &   2.47 \std{0.05} \\
\hline
\hline
Baseline: LSH~\cite{C02}  &  0.114 \std{0.0010} & 0.576 \std{0.0029} & 0.089 \std{0.0017} & 0.643 \std{0.0023} &   1.45 \std{0.05} \\
Baseline: ITQ~\cite{GL11} & 0.135 \std{0.0024} & 0.688 \std{0.0039} & 0.088 \std{0.0014} & 0.654 \std{0.0054} &   1.45 \std{0.05} \\
Baseline: PQ~\cite{JDS11} & 0.442 \std{0.0047} & 0.997 \std{0.0000} & 0.133 \std{0.0011} & 0.838 \std{0.0014} &   9.01 \std{0.01} \\
\hline
\end{tabular} ~ \hfill }
\smallskip
\caption{\label{tab:optimizations}
Analysis of polysemous codes (16 bytes/vector). 
The performance of \disidx does not depend on the index assignment. 
We give the performance of codes when compared in binary, 
before (\pq/\binary) and after (\polyd/\binary and \polyr/\binary) our optimization. 
Then we present the results for the proposed polysemous \hybrid strategy, 
which are almost as accurate as PQ while approaching the speed of binary methods. 
The Hamming thresholds are adjusted on the training sets so that the Hamming comparison filters out at least 95\% of the points.
The results are averaged over 5 runs, the sources of randomness being the k-means of the PQ training and the simulated annealing (the standard deviation over runs is always $<0.005$).
The last 3 rows are baselines provided for reference: LSH, ITQ and PQ. LSH uses a random rotation instead of random projection for better performance~\cite{BFJ14}.  
}
\end{table}

Table~\ref{tab:optimizations} details the performance of the aforementioned PQ constructions. First, note that the accuracy of \disidx is low, and that it is also relatively slow due to the lack of a dedicated machine instruction. Second, these results show that our index assignment optimization is very effective for improving the quality of the binary comparison. Without this optimization, the binary comparison is ineffective both to rank results (\pq/\binary), and to  filter (\pq/\hybrid). The ranking loss \polyr is slightly inferior to \polyd, so we adopt the latter in the following. 

\begin{figure}[t]
\begin{minipage}{0.52\linewidth}
\includegraphics[width=\linewidth]{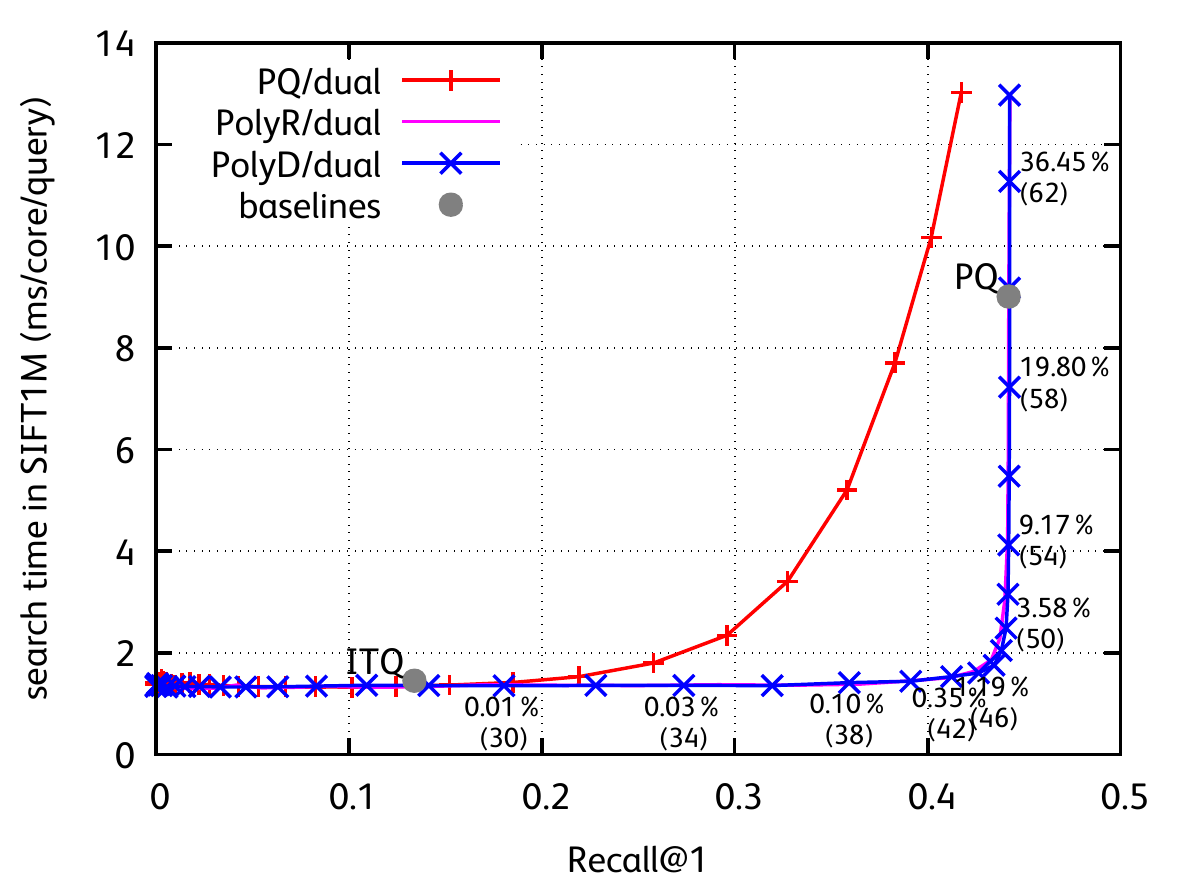}
\end{minipage} \hfill
\raisebox{4pt}{
\begin{minipage}{0.47\linewidth}
\caption{Impact of the threshold on the \hybrid strategy: Recall@1 vs search speed  for the \siftM dataset, with 128\,bits (16 subquantizers). The operating points for polysemous are parametrized by the Hamming threshold (in parenthesis), which influences the rate of points kept for PQ distance estimation. The tradeoffs obtained without polysemous optimization (\pq/\hybrid) and two  baselines (ITQ and PQ) are given for reference. 
\label{fig:hefilter}}
\end{minipage}}
\end{figure}

\begin{figure}[t]
\begin{minipage}{0.52\linewidth}
\includegraphics[width=\linewidth]{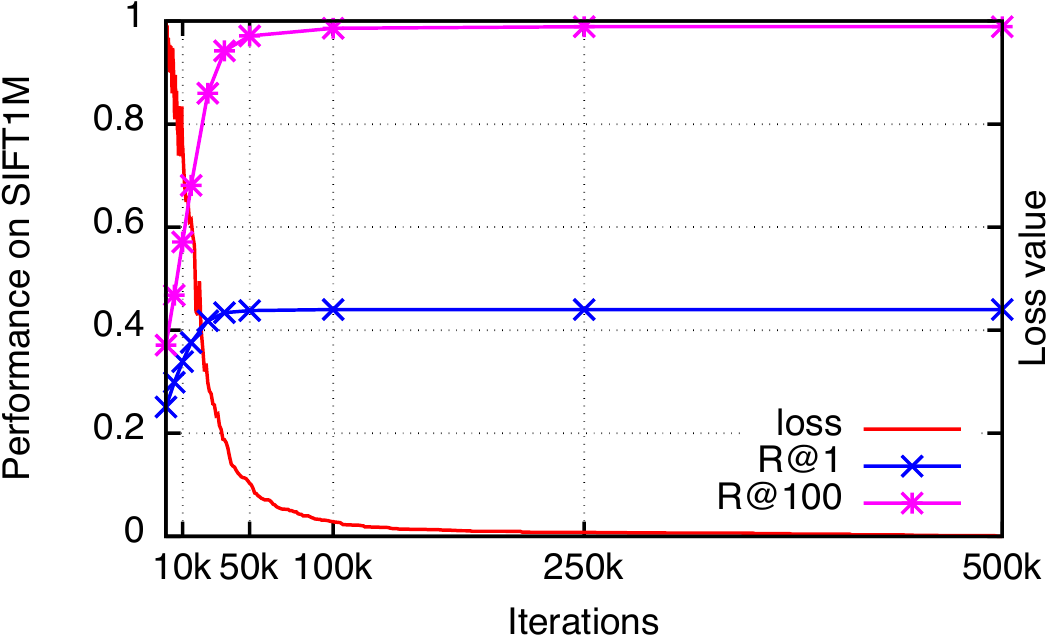}
\end{minipage}
\raisebox{6pt}{
\begin{minipage}{0.47\linewidth}
\caption{Performance of polysemous codes (\hybrid, $\tau=52$, 128 bits) along the iterations for the distance-based objective function described in Section~\ref{sec:polysemous} (results with the ranking loss are similar). Note that the initial state (0 iteration) corresponds to a product quantizer not yet optimized with our method. %
\label{fig:iterations}}
\end{minipage}}
\end{figure}

Figure~\ref{fig:hefilter} confirms the relevance of \polyd/\hybrid. 
It gives the performance achieved by this method for varying Hamming thresholds~$\tau$, which parametrizes the trade-off between speed and accuracy. %
Polysemous codes allow us to make almost no compromise: attaining the quality of \pq/\adc only requires a minor sacrifice in search time compared to binary codes. With threshold $\tau=54$, 90--95\% of the points are filtered out; for $\tau=42$ this raises to more than 99.5\%.
\smallskip
{\noindent \em Convergence:} Figure~\ref{fig:iterations} shows the performance of the binary filtering as a function of the number of iterations. 
The algorithm typically converges in a few hundred thousand iterations (1 iteration $=$ 1 test of possible index swaps). For a set of PQ subquantizer with 256 centroids each, this means a few seconds for the distance reconstruction loss \polyr and up to one hour for the ranking loss \polyr. %

\subsection{Comparison with the state of the art}

As mentioned in the related work Section~\ref{sec:related}, for large datasets the best trade-offs between accuracy, search time and memory are obtained~ by hybrid methods~\cite{JDS11,BL12} that combine a preliminary space partitioning, typically implemented through clustering, with compact codes learned on residual vectors. 
That is why we combine our polysemous codes with IMI~\cite{BL12}. This method partitions the space with a product quantizer (the ``coarse'' partitioning level) and uses PQ to encode residual error vectors. 
The search proceeds by selecting a few inverted lists at the coarse level, and then use the residual PQ codes to  estimate distances for the vectors associated with the selected lists. 
We further optimize the 
computation of the lookup tables involved in PQ when multiple lists are probed~\cite{BL14a}, and 
use an optimized rotation before encoding with PQ~\cite{GHKS13}.

Building upon this method, we learn polysemous codes for the residual PQ, which allows us to introduce an intermediate stage to filter out most of the list items, hence avoiding most of the distance estimation with PQ. 
Table~\ref{tab:bigann} gives a comparison against state-of-the-art algorithms on the \bigann dataset. We report both the timings reported for concurrent methods and our improved re-implementation of IMI. Note already that our system obtains very competitive results compared to the original IMI. Note, in the case where a single query vector is searched at a time, as opposed to batch mode, the coarse quantization becomes 50 to 60\% more expensive. 
Therefore, in the following we use %
$K=4096^2$ to target more aggressive operating points by reducing the fixed cost of the coarse quantizer. In this case, the results of \polyd/\hybrid gives a clear improvement compared to IMI$\star$ and the  state of the art. In particular, with 16 bytes we are able to achieve a recall@1=0.217 in less than 1\,ms on one core (0.38\,ms in single query mode, 0.64\,ms in batch mode). The binary filter divides by about 2$\times$ the search time, inducing only a small reduction of the Recall@1 score. 
\begin{table}[t]
\begin{tabular}{|lrr|rr|cc|rr|cc|}
\hline 
& code &  sizes $\rightarrow$ & \multicolumn{4}{c|}{8 bytes} & 
             \multicolumn{4}{c|}{16 bytes} \\
\hline
 & K\ \ \ \  & probes/cap & R@1 & R@100 & \multicolumn{2}{c|}{time\,(ms)} & R@1  & R@100 & \multicolumn{2}{c|}{time\,(ms)} \\
\hline
IMI~\cite{BL12} &  $16384^2$  & --/10k  & 0.158 & 0.706 & \multicolumn{2}{c|}{6}  & 0.304 & 0.740 & \multicolumn{2}{c|}{7} \\
IMI~\cite{BL12} &  $16384^2$  & --/30k  & 0.164 & 0.813 & \multicolumn{2}{c|}{13} & 0.328 & 0.885 & \multicolumn{2}{c|}{13} \\
\hline

IMI$\star$      & $16384^2$   & 1024/10k  & 0.159 & 0.719 &  1.57 & 2.58  & 0.313 & 0.753 & 1.92 & 2.89  \\
IMI$\star$      & $4096^2$    & 1024/10k  & 0.125 & 0.550 &  0.99 & 1.23  & 0.255 & 0.576 & 1.16 & 1.44  \\
IMI$\star$      & $4096^2$    & 16/10k    & 0.115 & 0.462 & 0.50 & 0.75 & 0.226 & 0.479 & 0.64 & 0.88 \\
\hline
IMI$\star$+\textsf{PolyD+ADC}  & $4096^2$    & 16/10k & 0.103 & 0.332 & \textbf{0.27} & 0.51 & 0.206 & 0.397 & \textbf{0.33} & 0.58 \\

\hline 
IMI$\star$      & $16384^2$   & 1024/30k   & 0.162 & 0.796 &  2.20 & 3.15  & 0.330  & 0.856 & 2.77 & 3.75  \\
IMI$\star$      & $4096^2$    & 1024/30k   & 0.134  & 0.696 & 1.35 & 1.61  & 0.295  & 0.755 & 1.77 & 2.07  \\
IMI$\star$      & $4096^2$    & 16/30k     & 0.117  & 0.505 & 0.59 & 0.81  & 0.238  & 0.532 & 0.75 & 1.01 \\
\hline
IMI$\star$+\textsf{PolyD+ADC}  & $4096^2$ & 16/30k & 0.106  & 0.370 & 0.33 & 0.56  & 0.217  & 0.447 & 0.38 & 0.64\\
\hline

\end{tabular} 
\smallskip%
\caption{Comparison against the state of the art on \bigann (1 billion vectors). 
We cap both the maximum number of visited lists and number of distance evaluations (column probes/cap).  
For the timings, using our improved implementation ($\star$), the first number is for queries performed in batch mode, while the second corresponds to a single query at a time. 
Our polysemous method is set to filter out 80\% of the codes. 
\tobeadded{*** redo with OPQ}
\label{tab:bigann}}
\end{table}

\begin{figure}[t]
\begin{minipage}{0.475\linewidth}
\includegraphics[width=\linewidth]{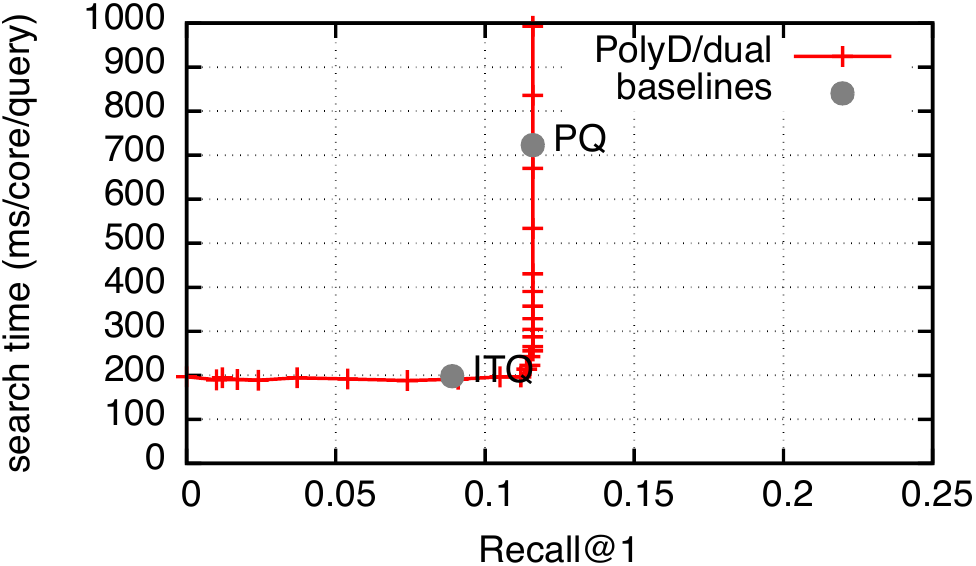}
\includegraphics[width=\linewidth]{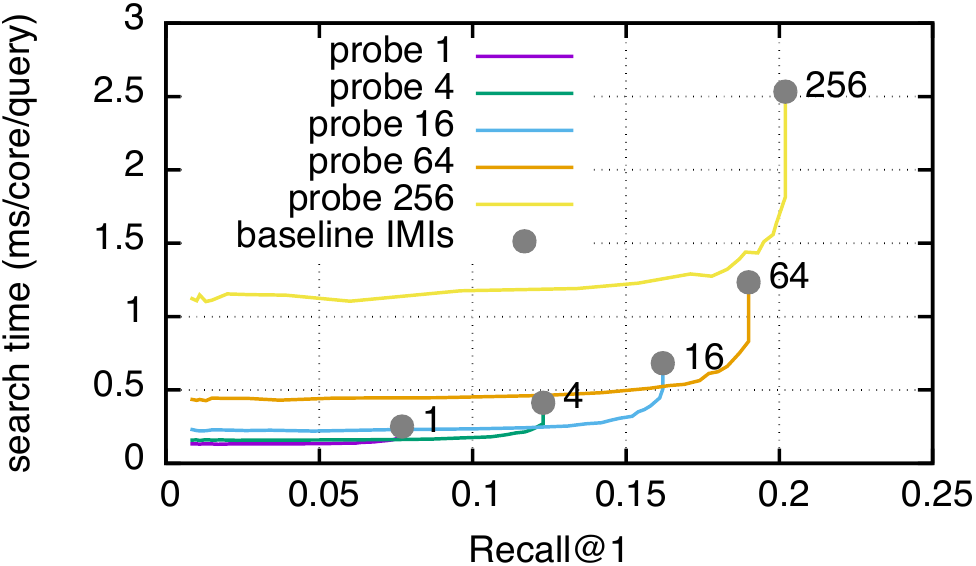}
\end{minipage}
\raisebox{10pt}{
\begin{minipage}{0.475\linewidth}
\caption{Performance on the \FYbig benchmarks. We use 20 bytes per vector (128 bits for the code and 4 bytes per identifier), \ie, per indexed image. \emph{Above:} For reference we give results obtained by methods that exhaustively compare the query to all vectors indexes based on their codes. As to be expected, the non-exhaustive methods (\emph{below}) achieve much better performance, especially when probing a large number of inverted lists (see ``probe 256''). Our proposal IMI+\polyd/\hybrid offers the best trade-off between memory, search time and accuracy by a fair margin. 
\tobeadded{*** update numbers}
\label{fig:YFbig}}
\end{minipage}}
\end{figure}

We now compare our method on the more challenging \FYbig benchmark, for which a single query amounts to searching an image in a collection containing 90 million images. 
Figure~\ref{fig:YFbig} reports the performance achieved by different methods. First observe that the non-exhaustive methods (\emph{bottom}) are at least 2 orders of magnitude faster than the methods that compare the codes exhaustively (\emph{top}), like ITQ. The former are able to find similar images in a few seconds. 
Again, our polysemous strategy IMI+\polyd/\hybrid offers a competitive advantage over its competitor IMI. Our method is approximately $1.5\times$ faster for a negligible loss in accuracy. 

%% file: graph.tex
\section{Application: large-scale k-NN image graph}
\label{sec:graph}

\newcommand{\knnigquery}[1]{\includegraphics[width=1.5cm]{figs/knngraph/#1} \hfill \raisebox{0.75cm}{$\boldsymbol{\rightarrow}$} \hfill}
\newcommand{\knnig}[1]{\includegraphics[width=1.5cm]{figs/knngraph/#1}}

As an application to our fast indexing scheme, 
we now consider the problem of building the approximate k-NN graph of a very large image collection.
For this experiment, we make use of the 95,063,295 images available in the Flickr 100M dataset.
As was the case in Section~\ref{sec:experiments}, we use 4,096D AlexNet features reduced to 256D with PCA.
For the graph construction, we simply compute the k-NN with $k$=100 for each image in turn.
This takes 7h44 using 20 threads of a CPU server. Note that the collection that we consider is significantly larger than the ones considered in previous works~\cite{DCL11,WJGZGL12} on kNN graph. Moreover, our approach may be complementary with the method proposed by Dong \etal ~\cite{DCL11}. 

\begin{figure}[th!]
\input{knngraph_images.tex}
\vspace{-8pt}
\caption{Examples of image modes and their neighbors in the graph. For each reference image (left), we show the corresponding image neighbors in the kNN graph on its right. 
\label{fig:img_results}}
\vspace{-12pt}
\end{figure}
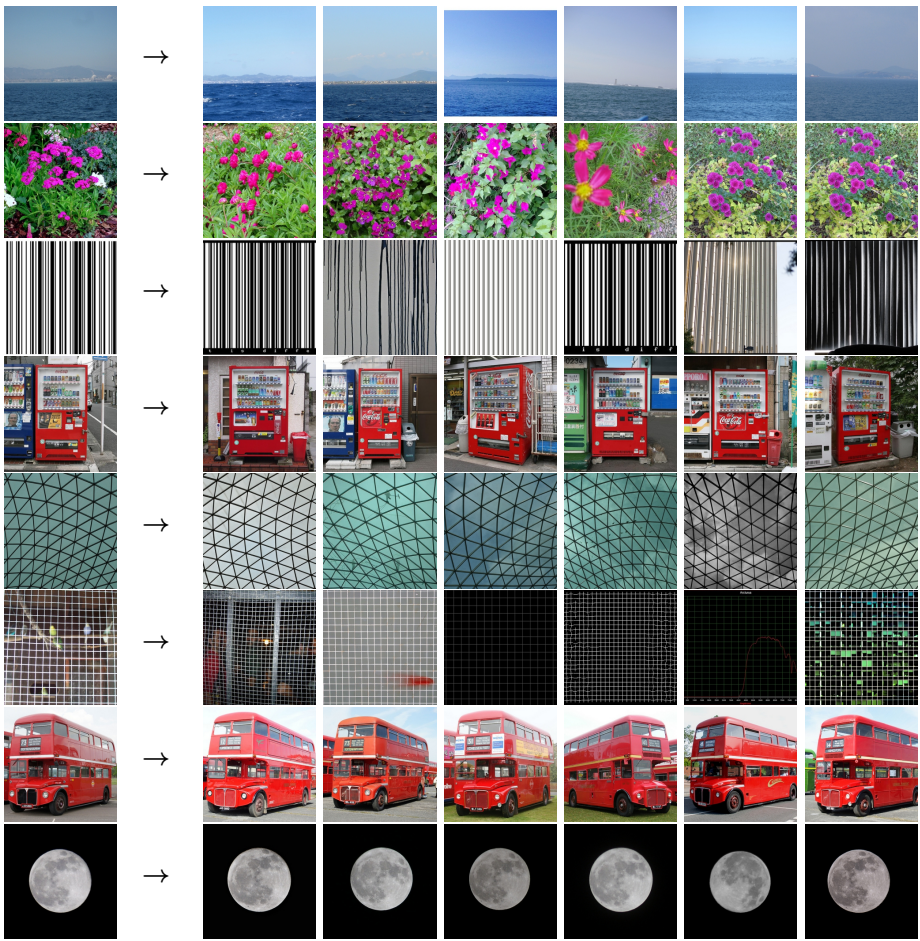

For visualization purposes, we seek the modes following a random walk technique~\cite{CL12}:
we first iteratively compute the stationary distribution of the walk,
(\ie the probability of each node to be visited during the random walk)
and then consider as modes each local maximum of the stationary probability in the graph.
We find on the order of 3,000 such maxima.
Figure~\ref{fig:img_results} depicts a sample of these maxima as well as their closest neighbors.
We believe that these results are representative of the typical quality of the found neighbors, except that, for privacy reasons, we do not show the numerous modes corresponding to faces, 
of which we found many including specialized modes of ``pairs of persons'', ``clusters of more than two persons'' or ``baby faces''.

\vfill

%% file: knngraph_images.tex
\knnigquery{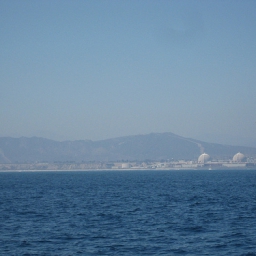}
\knnig{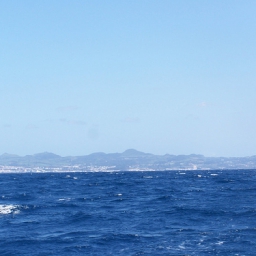}
\knnig{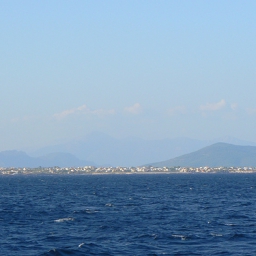}
\knnig{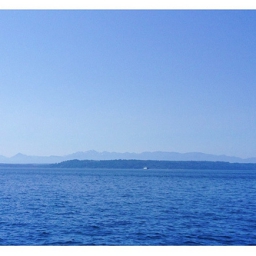}
\knnig{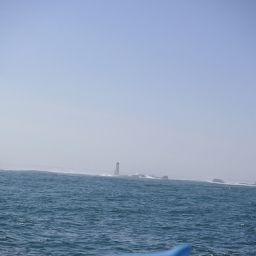}
\knnig{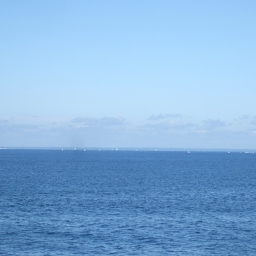}
\knnig{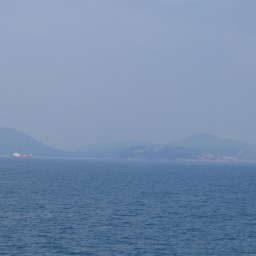}
\\
\knnigquery{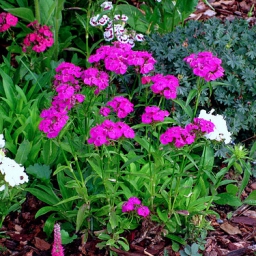}
\knnig{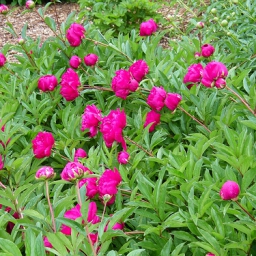}
\knnig{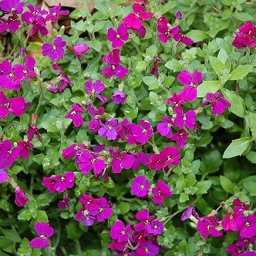}
\knnig{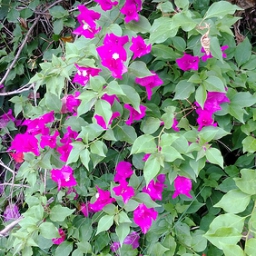}
\knnig{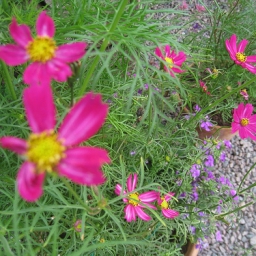}
\knnig{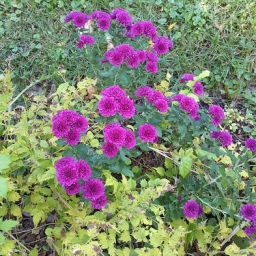}
\knnig{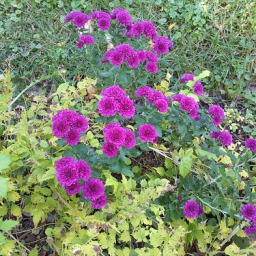}
\\
\knnigquery{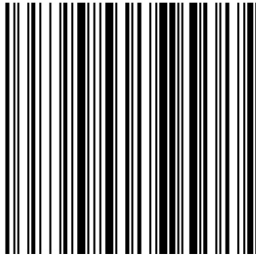}
\knnig{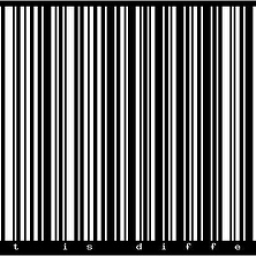}
\knnig{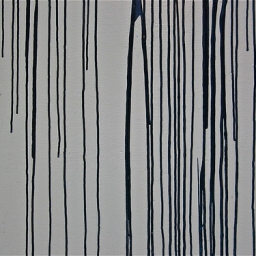}
\knnig{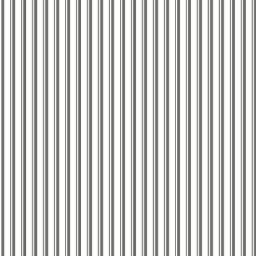}
\knnig{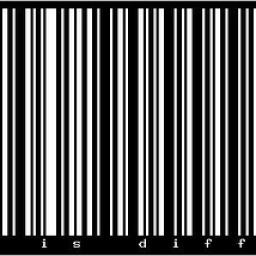}
\knnig{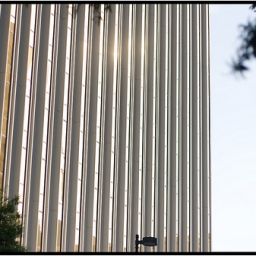}
\knnig{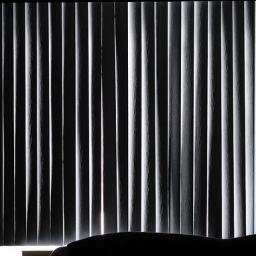}
\\
\knnigquery{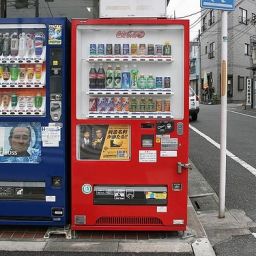}
\knnig{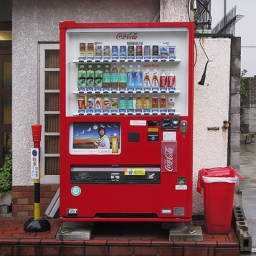}
\knnig{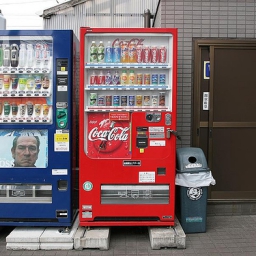}
\knnig{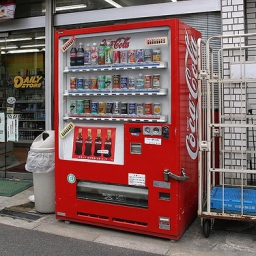}
\knnig{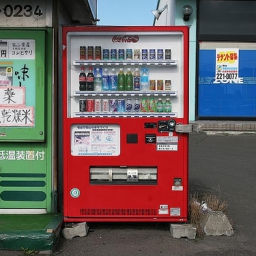}
\knnig{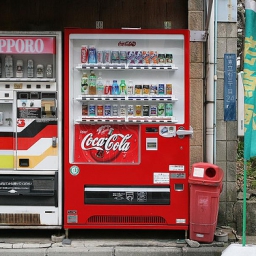}
\knnig{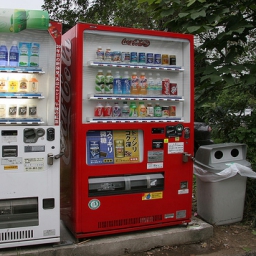}
\\
\knnigquery{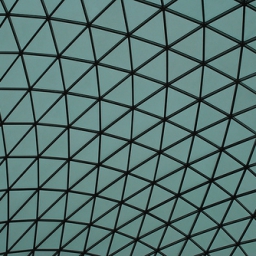}
\knnig{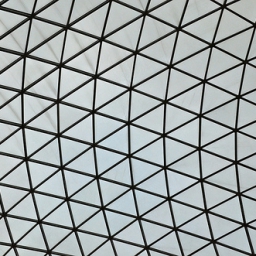}
\knnig{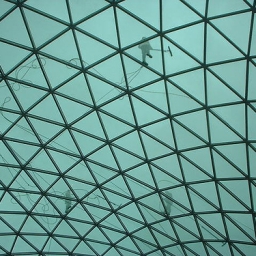}
\knnig{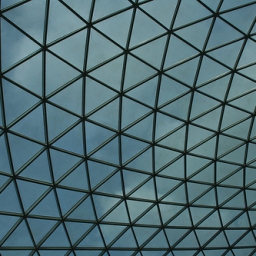}
\knnig{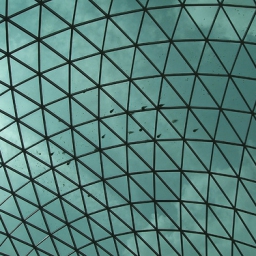}
\knnig{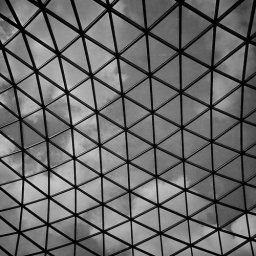}
\knnig{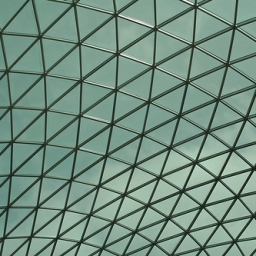}
\\
\knnigquery{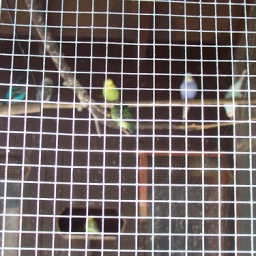}
\knnig{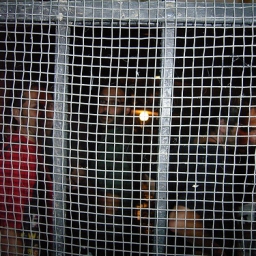}
\knnig{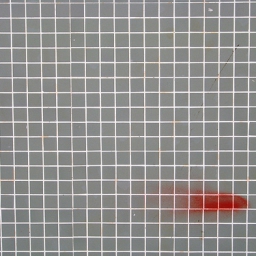}
\knnig{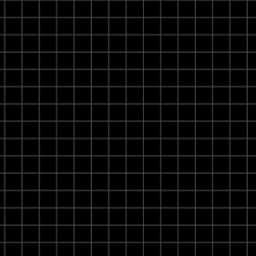}
\knnig{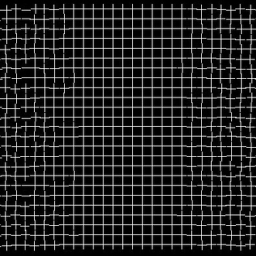}
\knnig{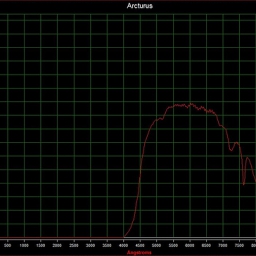}
\knnig{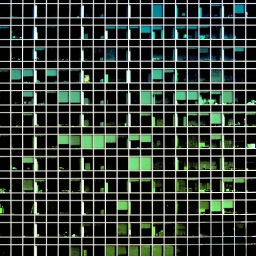}
\knnigquery{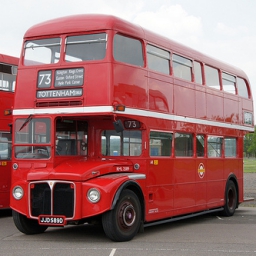}
\knnig{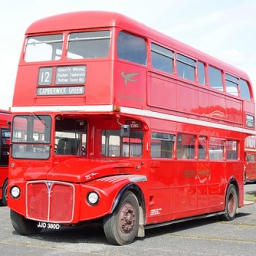}
\knnig{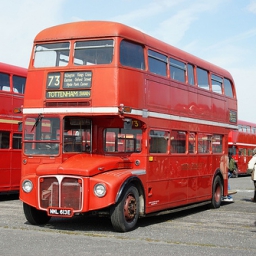}
\knnig{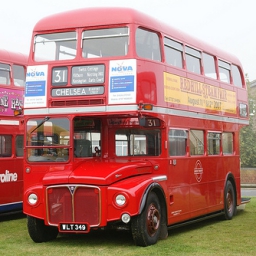}
\knnig{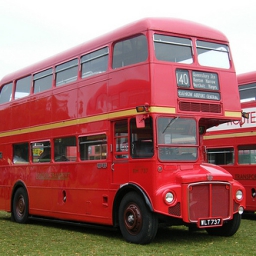}
\knnig{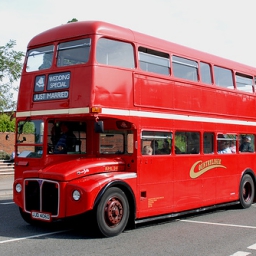}
\knnig{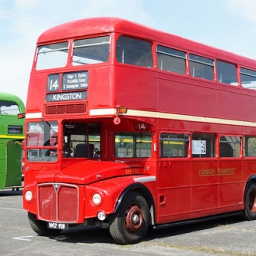}
\\
\knnigquery{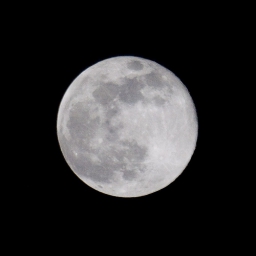}
\knnig{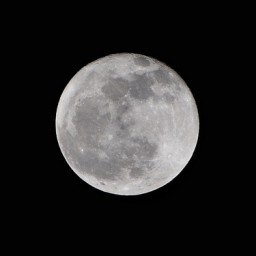}
\knnig{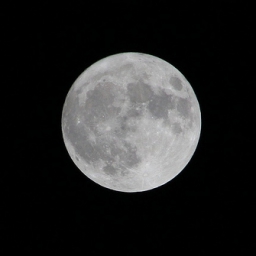}
\knnig{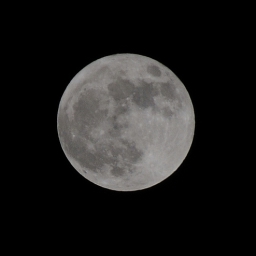}
\knnig{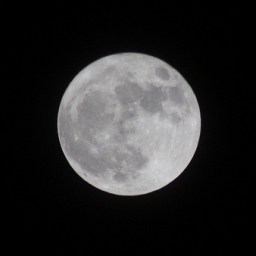}
\knnig{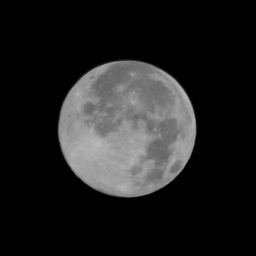}
\knnig{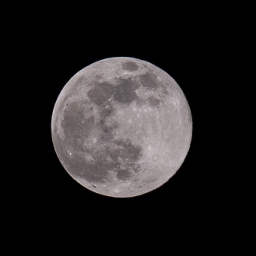}

%% file: conclusion.tex
\section{Conclusion}
\vspace{-2mm}
In this work, we introduced polysemous codes,
\ie, codes that can be interpreted both as binary codes and as product quantization codes.
These complementary views are exploited for very large-scale indexing in a simple two-stage process that first involves filtering-out 
the majority of irrelevant indexes using the fast Hamming distance on binary codes,
and then re-ordering the short list of candidates using the more precise but also slower asymmetric distance on PQ codes.
This yields a competitive indexing scheme that combines the best of both worlds:
its speed is comparable to that of pure binary techniques and its accuracy matches that of PQ.

\subsubsection*{Acknowledgements.}
We are very grateful to Armand Joulin and Laurens van de Maaten for providing the Flicrk100M images
and their CNN descriptors. Alexandre Sablayrolles had the idea of extending the OPQ method to reduce 
the number of dimensions.

\section*{Additional results}

We present results obtained after the ECCV final version submission. We compare operating points with recent papers~\cite{WieschollekCVPR16,BL16}. The only addition to our method is to reduce the dimension of the descriptors before encoding it, within the OPQ~\cite{GHKS13} transformation. The runtime parameters (multi-probe, Hamming threshold) are tuned automatically to optimize the tradeoff between speed and accuracty. 

Table~\ref{tab:newres} compares with a recent hybrid CPU/GPU method~\cite{WieschollekCVPR16}, for a given accuracy. It shows that our implementation is slower with a single thread, but is much faster when all cores are used. %

Deep1B is a 1-billion CNN vector dataset, introduced by Babenko et al~\cite{BL16}. The vectors are reduced to 96 dimensions, otherwise the dataset statistics are the same as for SIFT1B. We compare against the method that was introduced along with the dataset, allowing~20-byte codes to have a similar memory usage. Table~\ref{tab:newres} shows that our method is about 5 times faster for a comparable accuracy.

\newcommand{\idn}[1]{#1}

\begin{table}
\begin{center}

\begin{tabular}{|ll|rr|}
\hline
\multicolumn{4}{|c|}{SIFT1B (8 bytes per code)}\\
\hline
method                & hardware & 1-R@10 & time (ms) \\
\hline
Wieschollek~\cite{WieschollekCVPR16}             & Titan X  & {\em 0.35}    & {\em 0.15} \\
\idn{OPQ8\_64,IMI2x13,PQ8} & 1 thread & 0.349 & 0.485 \\
                      &20 threads& 0.349 & 0.035  \\
\hline
\end{tabular}

\begin{tabular}{|ll|rr|}
\hline 
\multicolumn{4}{|c|}{Deep1B (20 bytes per code)}\\
\hline
method                & hardware & 1-R@1 & time (ms) \\
\hline
Babenko~\cite{BL16}                   & 1 thread  & {\em 0.45} & {\em 20} \\
\idn{OPQ20\_80,IMI2x14,PQ20} & 1 thread & 0.456 & {\bf 3.66} \\
\hline
\end{tabular}

\end{center}
\caption{\label{tab:newres}
Additional results on the datasets SIFT1B and Deep1B~\cite{BL16}. Parameters of the methods: OPQ8\_64 = OPQ~\cite{GHKS13} to 8 blocks, and reduction to 64\,D, IMI2x13 means that we use an Inverted multi-index~\cite{BL12} with 2 sub-quantizers, each having $2^{13}$ centroids. PQ8 refers to a product quantizer with 8-byte codes.
}
\end{table}

\newpage